\documentclass{WileyMSP-template}

\usepackage{xr}
\usepackage{xcolor}
\usepackage{amsmath}
\usepackage{textcomp}

\begin{document}

\pagestyle{fancy}
\rhead{\includegraphics[width=2.5cm]{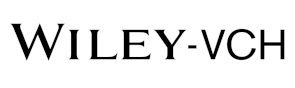}}

\title{Responsive Hydrogel-based Modular Microrobots for Multi-functional Micromanipulation}

\maketitle


\author{Liyuan Tan*}
\author{David J. Cappelleri}


\dedication{}

\begin{affiliations}
L. Tan, Prof. D. J. Cappelleri\\
School of Mechanical Engineering\\
Purdue University\\
West Lafayette, IN 47907, USA\\
Email Address: tan328@purdue.edu

Prof. D. J. Cappelleri\\
Weldon School of Biomedical Engineering\\
Purdue University\\
West Lafayette, IN 47907, USA

\end{affiliations}



\begin{abstract}

Microrobots show great potential in biomedical applications such as drug delivery and cell manipulations. However, current microrobots are mostly fabricated as a single entity and type and the tasks they can perform are limited. In this paper, modular microrobots, with an overall size of 120 $\micro$m $\times$ 200 $\micro$m, are proposed with responsive mating components, made from stimuli-responsive hydrogels, and application specific end-effectors for microassembly tasks. The modular microrobots are fabricated based on photolithography and two-photon polymerization together or separately. Two types of modular microrobots are created based on the location of the responsive mating component. The first type of modular microrobot has a mating component that can shrink upon stimulation while the second type has a double bilayer structure that can realize an open and close motion. The exchange of end-effectors with an identical actuation base is demonstrated for both types of microrobots. Finally, different manipulation tasks are performed with different types of end-effectors. 

\end{abstract}


\section{Introduction}

Untethered microrobots have been found promising for biomedical applications, such as drug delivery \cite{Erkoc2019,Lee2021MagneticallyAbility,Jang2019TargetedReview}, single cell/tissue manipulation \cite{Jeon2019,Tan2019-PIVMicroswimmers}, and single-cell studies \cite{Grexa2020Single-cellMicrorobot}. Other potential applications like micromanipulation of micro objects for low-cost manufacturing are also popular \cite{Johnson2020DesignTasks}. In the past decade, many solutions have emerged for micromanipulation tasks with microrobots. For example, helical microswimmers have been fabricated for cargo transport under a rotating magnetic field \cite{Tottori2012MagneticTransport}. Force-sensing microrobots are achieved for safe manipulation with controlled force using a gradient magnetic field \cite{Jing2019,Adam2020StiffnessMicrorobot}.

In the past few years, the rapid development of smart polymers, such as hydrogels and liquid crystalline polymers, have introduced new design scenarios for microrobots \cite{Tan2021SmartMachines,Adam20214DApplications}. For example, microcrawlers \cite{Rehor2020PhotoresponsiveActuation}, helical microswimmers \cite{Huang2016,Huang2017b}, and microgrippers \cite{Jia2019UniversalMicrogripper,Jin2020UntetheredBiopsy,Ongaro2016} have been fabricated with hydrogels. The liquid crystalline polymers have also been used to achieve microcrawlers with feet \cite{Zeng2015Light-FueledWalkers}. These smart polymers are able to deform after fabrication under stimuli, such as temperature, light, and chemical changes, which provides an additional freedom of design in addition to geometry \cite{Tan2021SmartMachines,Soto2021SmartMicrorobots}. The material response allows for the introduction of advanced functionalities. Compared with microrobots obtained from traditional stiff materials, hydrogels are soft and responsive so that they can be used to achieve microswimmers with adaptive locomotion, either passive or active \cite{Huang2019,Tan2021ModelingLocomotion}. 

Fabrication of modular microrobots that have multiple micro-components achieving different tasks is one of the targeted goals in this area \cite{Sitti2015BiomedicalMilli/Microrobots,Tsang2020RoadsMicroswimmers}. Achieving modular microrobots from sub-components could lead to more advanced and optimal microrobots \cite{Sitti2015BiomedicalMilli/Microrobots,Sitti2020ProsMicrorobots}. However, it was said that the realization of these modular microrobots is restricted due to the lack of materials and techniques required to solve the sophisticated design of microrobots \cite{Ceylan2019TranslationalMicrorobots,Zheng2022ProgrammableMicrorobots}. Limited by the stiffness of the materials previously used for microrobot fabrications, modular microrobots that can maintain both their individual and assembled modes without continuous stimulation are hard to achieve. Cheang et al. achieved modular microswimmers using the magnetic dipole-dipole interaction between microbead subunits \cite{Cheang2016VersatileSubunits}. However, the assembled states vary from one another because of the inconsistent magnetic moment of the base module as well as the subunits. Moreover, since both the base module and the subunits are actuated under a global field, the assembly process is challenging because their swimming states must be controlled in a relative manner. Pauer et al. also developed modular microswimmers with patterned subunits \cite{Pauer2021ProgrammableMicroswimmers}. Those subunits assembled into the final microswimmers through the magnetic dipole-dipole interactions. The assembled microswimmers can also aggregate into a larger one as a result of the interaction. However, the reversibility of the assembly has not been discussed. Recently, modular microrobots have been developed for complex assemblies achieving gear structures, snake-shaped structures, and micro-vehicles \cite{Dai2020IntegratedMicrorobots}. The locomotion and the assembly process are achieved by lifting the parts with a bubble generated optothermally. However, the locomotion of the microrobots is realized through an inefficient approach as the bubble is generated discretely. Recently, Alapan et al. fabricated microstructures with assembly capabilities by taking advantage of dielectrophoretic forces generated by specially designed structural features under an electric field \cite{Alapan2019Shape-encodedMicromachines}. These non-magnetic microstructures are able to move under magnetic fields when they are incorporated with magnetic spheres which are assembled into the microstructures via the dielectrophoretic forces.

By taking advantage of the responsiveness of hydrogels, that deform upon stimulation, self-assembly with multiple subunits has been achieved. For example, self-assembly of specially designed circular/rectangular discs at the air/water interface is achieved in \cite{Kim2019Light-DrivenSurfers,Bae2017ProgrammableMultipoles}. The attraction forces are induced by photothermal deformations of the hydrogel discs. The locomotion of the discs can be obtained by moving a light between two discs. However, these subunits can only work at the air/water interface and the assembled structures are not able to move except for the case with only two subunits and a moving light. A recent study also uses hydrogel to achieve a modular microrobot by making different functional components at the same time by an aniso-electrodeposition method~\cite{Zheng2022ProgrammableMicrorobots}. This method provides a one-step fabrication process for modular microrobots that can perform propulsion, grasping, and object delivery. However, these microrobots still cannot exchange modules and a new microrobot is required once the gripping component is dissolved after delivery. 

Even though there is a growing number of investigations on hydrogel-based microrobots, modular microrobots with orthogonal actuation over the locomotion and the assembly process under a gradient magnetic field have not been explored. The number of reports on modular microrobots is also limited. Some applications of assembly only achieved the assembly process itself, not achieving any advanced functionalities by using different modules of the modular microrobots. 

In this paper, we fabricate modular microrobots using the method of photolithography and two-photon polymerization (TPP) with footprint of 120 $\micro$m $\times$ 200 $\micro$m. The modular microrobots contain two parts: an end-effector and a magnetic base. The end-effector and the base are connected by a responsive mating component (RMC). Depending on the location of the RMC, the modular microrobot can be categorized into two types: the first has a RMC (single-layer design) as part of the magnetic base while the second one has an RMC (bialyer-design) with the end-effector. The RMCs show a significant size shrinkage in water and have a size slightly larger than designed in an environment containing 40\% of water and 60\% of ethyl lactate (EL). Both types of modular microrobots are investigated and validated for the mating process. Moreover, the proposed modular microrobots are used to achieve different micromanipulation tasks. 

\section{Results and Discussions}

\subsection{Design Overview}

\begin{figure}[t!]%
\centering
\includegraphics[width=0.8\linewidth]{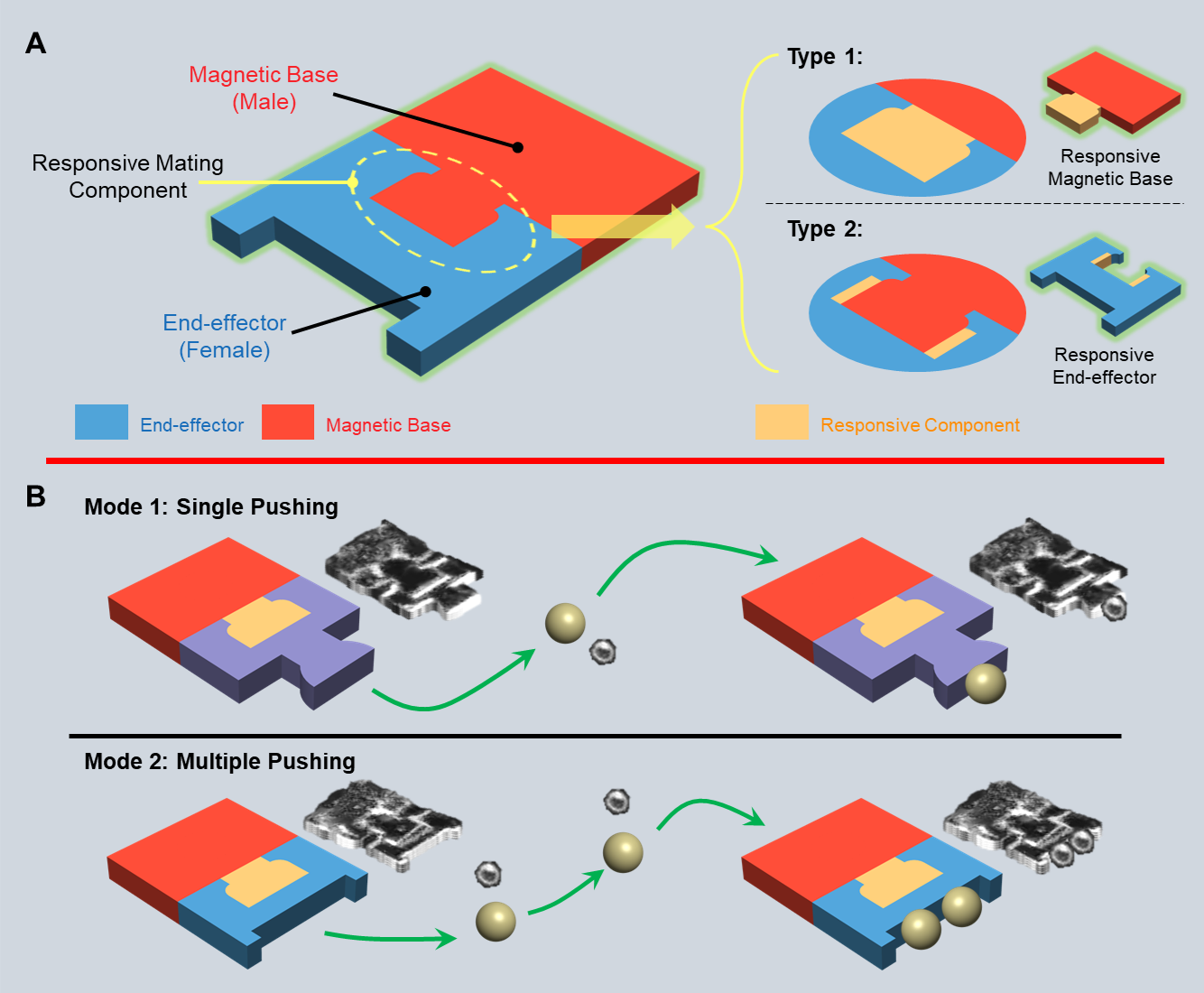}
\caption{Schematic of the modular microrobots connected by an RMC. (A) General design of modular microrobots with male and female features. The location of the RMC defines the microrobot types: Type 1 = base; Type 2 = end-effector. (B) Different modes of modular microrobots achieved with different types of end-effectors. }\label{schematic}
\end{figure}

The modular microrobots are designed with a male-female connection between different modules as shown in \textbf{Figure \ref{schematic}}A. In this paper, the male feature is placed with the magnetic base while the female feature is on the end-effector. For the proposed modular microrobots, the attachment and detachment of different modules are achieved via the RMC where the male and female parts are connected as illustrated in the yellow circular region. Based on where the RMC is, the designs can be separated into two types. The RMC of the Type 1 microrobot is fabricated with the magnetic base, while the RMC for Type 2 is made with the end-effectors. The RMC of the Type 1 microrobot is the mail feature that is able to shrink upon stimulation and in order to detach from the end-effectors. For Type 2, the RMS is part of the female feature of the end-effector and is designed with a double-bilayer structure. With this structure, the RMS can perform an opening and closing motion in different environments. \textbf{Figure \ref{schematic}}B demonstrates different modes that can be realized by the modular microrobots. Various applications can be done using different end-effectors with an identical magnetic base. For example, pushing with only one sphere is obtained with a tipped end-effector (Mode 1) while with an end-effector having a larger opening the microrobot can push multiple spheres (Mode 2).  

\subsection{Simulations of the Responsive Mating Component}

In order to understand the working mechanism of the RMCs, simulations based on finite element analysis (FEA) are performed for qualitative studies of the proposed modular microrobots. The simulations are achieved using commercial FEA software ABAQUS 2018. In these simulations, the non-responsive parts are simulated as neo-hookean materials with $C_{10}$ = 0.015 MPa and $D_1$ = 10 MPa$^{-1}$ (properties used to define the neo-hookean materials in ABAQUS considering compressibility) while the responsive parts are implemented using the user-defined hyperelastic material subroutine UHYPER for environmental responsiveness. The final form of the free energy function used for the UHYPER implementation is given as:

\begin{equation}\label{eq:Helmholtz2}
\begin{split}
    W &= \frac{1}{2}Nv[\lambda_0^{-1}{J^\prime}^{\frac{2}{3}}\overline{I_1}-3\lambda_0^{-3}-2\lambda_0^{-1}\ln{(\lambda_0^{3}J^\prime)}]\\
    &+(J^\prime-\lambda_0^{-3})\ln{(\frac{J^\prime}{J^\prime-\lambda_0^{-3}})}-\frac{\chi}{\lambda_0^{6}J^\prime}-\frac{\mu}{kT}(J^\prime-\lambda_0^{-3})
\end{split}
\end{equation}

\noindent after introducing a Legendre transform and using the free-swelling state as the reference state \cite{Tan2021ModelingLocomotion} where 
$J$ and $J^\prime$ are the volume ratios to the dry and free-swelling states, respectively, and $J=\det{\mathbf{F}}=\lambda_0^3J^\prime$ and $\overline{I_1}$ is the invariant of the deformation gradient $\mathbf{F}$. 
$Nv$ is a dimensionless parameter representing the number of polymer chains at the dry state, $\chi$ is the interaction parameter characterizing the interactions between the polymer matrix and the solvent, and $\mu$ is the chemical potential between the polymer matrix and the solvent before the system reaches equilibrium. The material properties used for the simulation are used with $Nv$ = 0.0854, $\lambda_0$ = 2.2617, and $\chi$ = -0.7363, respectively. Discussions of the simulation results can be found in the following sections accompanied by the corresponding designs of different mating components.  

\begin{figure}[t!]%
\centering
\includegraphics[width=0.8\linewidth]{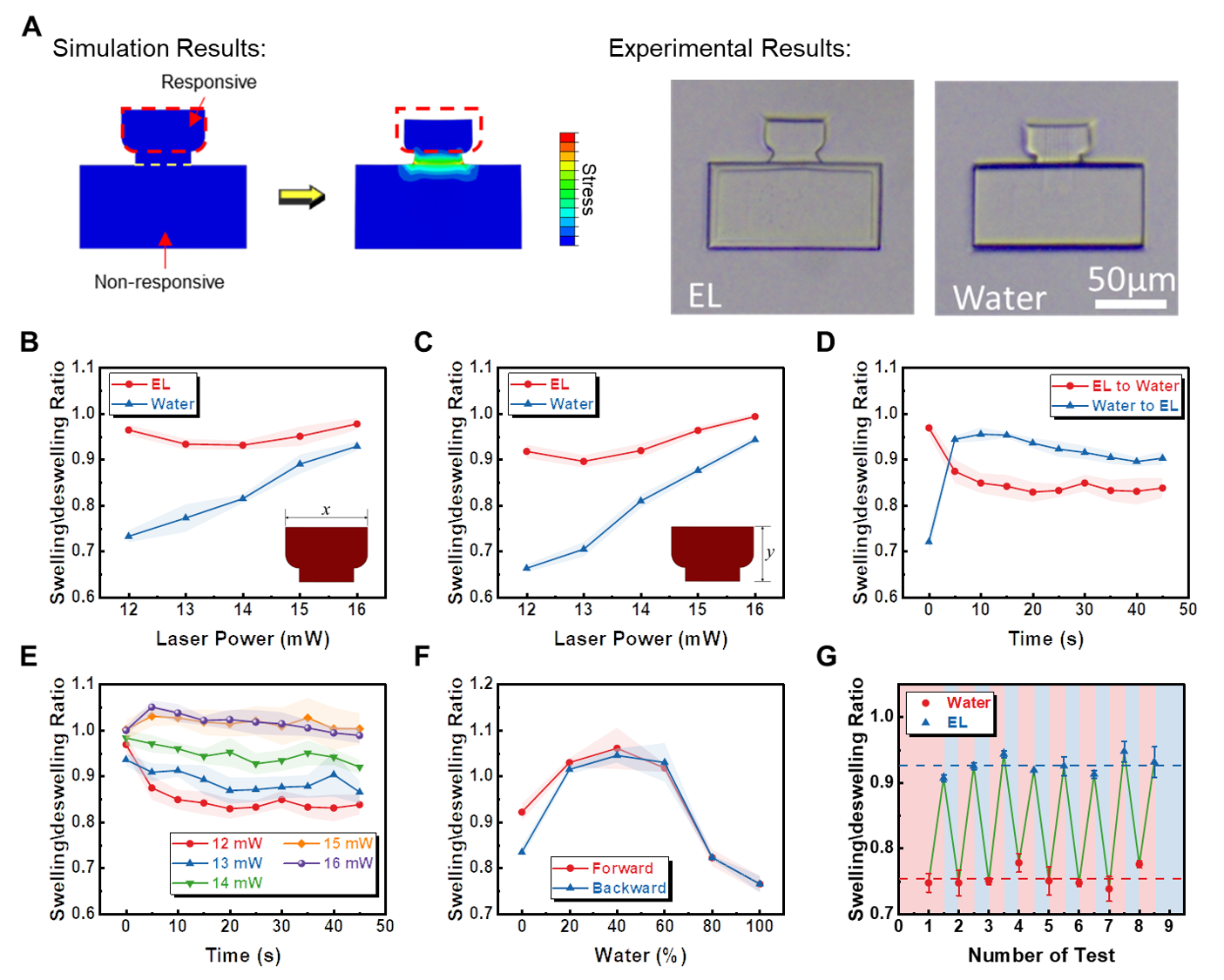}
\caption{Swelling behaviors of the responsive mating designs with the Type 1 design. (A) Qualitative simulation of deswelling of the mating component. (B) Swelling ratio of the responsive mating designs with different printing LP along the $x$-direction at equilibrium states with five samples ($n$ = 5). (C) Swelling ratio of the responsive mating designs with different printing LP along the $y$-direction at equilibrium states with $n$ = 5. The data in (B) and (C) are collected right after the structures are printed and well-developed for minimal residue. (D) Transition swelling of the responsive mating design (LP = 13 mW) from EL to IPA/water and the corresponding reverse behaviors with $n$ = 5. (E) Transition swelling of the responsive mating designs with different LPs from EL to water with $n$ = 5. (F) Deswelling of the printed responsive mating design with an LP = 12 mW in different chemical environments (different compositions of water and EL) with $n$ = 5. (G) Repeatability of shape (deswelling ratio) change of the mating component between 100\% water and EL (0\% water) with $n$ = 5. Error bars in (B) to (F) are represented with shaded areas with standard deviations (SD).}\label{fig_modular1}
\end{figure}

\subsection{Fabrication and Validation of the Responsive Mating Components}

The RMCs are fabricated using a commercial TPP system (Photonic Professional GT2, Nanoscribe GmbH) using a 63$\times$ objective with a poly-(\emph{N}-isopropylacrylamide) (PNIPAM) hydrogel material. The hydrogel precursor for printing is prepared by mixing chemicals in two steps as described in \cite{Tan2023DesignHAMMR}. In brief, 1.6 g of \emph{N}-isopropylacrylamide (NIPAM), 0.8 mL of acrylic acid (AAc), and 0.15 g polyvinylpyrrolidone (PVP) are dissolved in 1 mL of EL followed by vigorous stirring for complete dissolution. Then 2.5 mL of the solution obtained above is mixed with 0.4 mL of dipentaerythritol pentaacrylate (DPEPA), 0.5 mL of triethanolamine (TEA), and 100 $\mu$L of 4,4’-bis(diethylamino)benzophenone/\emph{N},\emph{N}-dimethylformamide (EMK/DMF) solution (1 to 4 weight ratio). Complete dissolution is achieved by mixing with a magnetic stirrer. During the hydrogel printing, the oil-immersion mode is adopted with a droplet of the precursor applied on a circular coverslip with a diameter of 30 mm. Immersion oil is applied between the coverslip and the objective. The swelling and deswelling ratios of the hydrogels are defined as the ratio of the resulting stimulated size to the designed size. If the resulting stimulated size is larger than the designed size, a swelling ratio is used, and vice versa for the deswelling ratio. 

Using the finite element simulation, a qualitative validation is first performed to prove the feasibility of the RMC for the Type 1 microrobot as shown in \textbf{Figure \ref{fig_modular1}}A. As can be seen from the result, the inner stress within the RMC (red dash region) is at a low level maintaining minimal energy. However, the connecting region between the responsive and non-responsive materials presents a high stress that may cause delamination if the two materials are not compatible. \textbf{Figure \ref{fig_modular1}}A also provides experimental results of a printed sample deforming in different environments showing good agreement with the simulation. To gain a deeper understanding of the RMC, various samples are printed and tested with different laser powers (LPs), which result in different crosslinking densities, printed with a non-magnetic base. All RMCs are printed with a laser scanning speed (SS) of 8 mm/s. The LP and SS together contribute to the overall dose polymerizing the hydrogel. As can be seen in \textbf{Figure \ref{fig_modular1}}B and \textbf{Figure 
 \ref{fig_modular1}}C, the designs printed with different LPs maintained close to the designed sizes (60 $\micro$m and 40 $\micro$m for $x$ and $y$ directions, respectively) with swelling ratios all above 95\%. However, there is a slight drop as the LP is decreased. The swelling ratio has a significant drop along the $x$ direction when compared against the $y$ direction when the LP becomes smaller. For samples with an LP of 12 mW, the swelling ratio along the $x$ direction in IPA is even slightly lower than in water. Samples in water generally have a lower swelling ratio for all the tested LPs. However, by increasing the LP you can get highly crosslinked structures having a swelling ratio of $\sim$1 in water. 

The transition swelling behaviors of a sample printed with an LP of 12 mW between EL and water are also studied, for both the forward and backward swelling, as shown in \textbf{Figure \ref{fig_modular1}}D. The swelling/deswelling from water to EL is fast within 5s to a relatively stable state. However, the transition from EL to water is much slower. The deswelling ratio is expected to be at around the equilibrium state while it only reaches 0.838 after 45 seconds. The speed of deswelling decreased as the chemical potential $\mu$ decreased. Stable data are measured after an hour. However, the the transition measurement could be delayed due to the adhesion between the sample and the substrate. \textbf{Figure \ref{fig_modular1}}E shows that the transition speed of deswelling is increased with LPs. \textbf{Figure \ref{fig_modular1}}F shows a sample printed with 12 mW of LP in different solvents (with different compositions of water and EL) at the equilibrium state. For convience, different compositions of water and EL is referred to as percentage water, e.g., 40\% water is a mixture of 40\% of water and 60\% of EL. As depicted in the figure, a peak is found at around 40\% of water showing a swelling effect with a ratio higher than 1. The repeatability of the swelling/deswelling of the responsive part is tested for 8 cycles as shown in \textbf{Figure \ref{fig_modular1}}G. For 8 cycles, the average deswelling ratio is 0.927 in pure EL and 0.753 in water. The results show a good consistency for repeatable use.  

\begin{figure}[h]%
\centering
\includegraphics[width=0.8\linewidth]{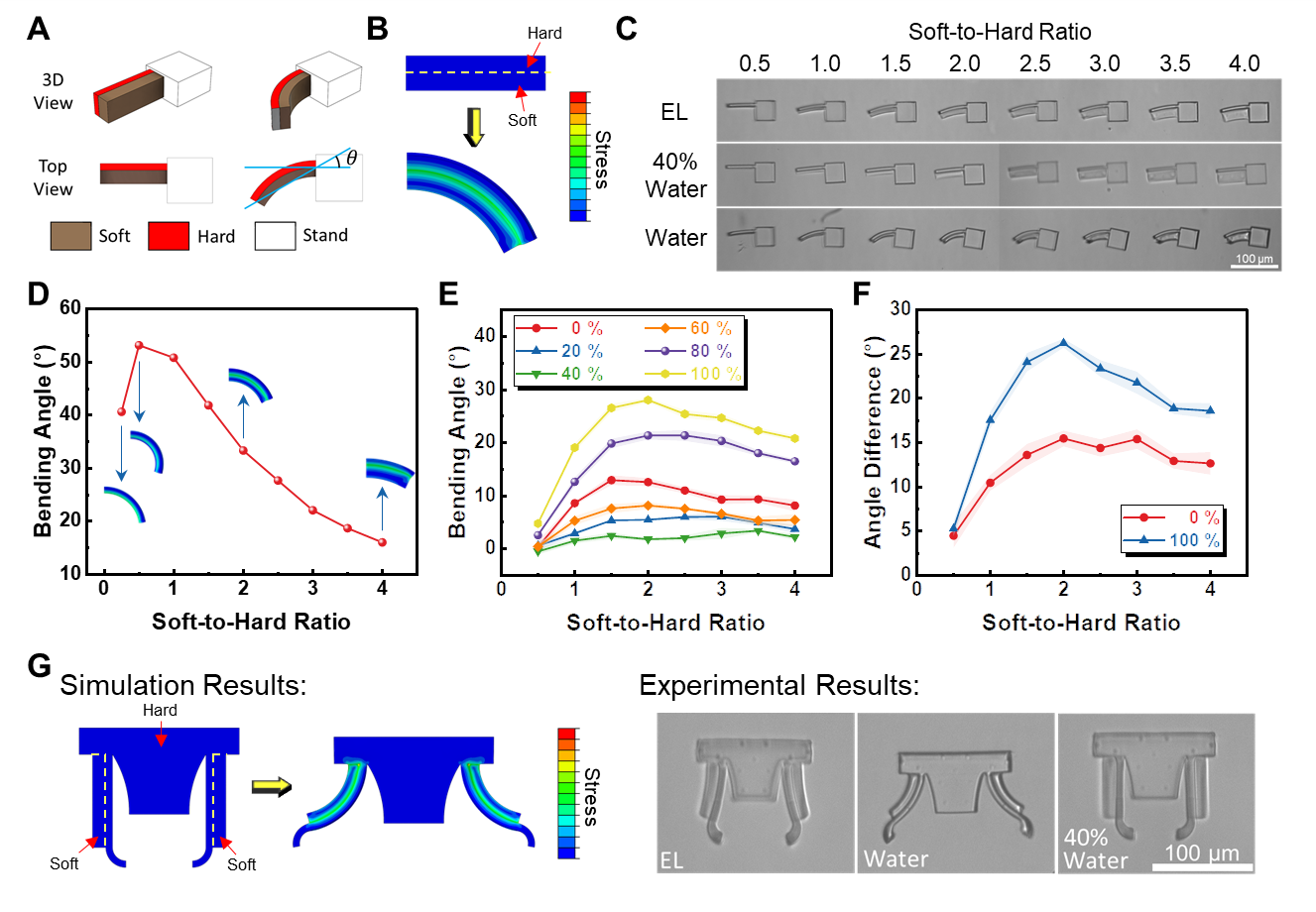}
\caption{Bending behavior of bilayer structures with different thickness ratios. (A) Systematic of the suspended bilayer structures. (B) Qualitative simulation of a bilayer structure. (C) Images of bilayer structures in environments with different EL/water compositions. (D) Bending angles measured based on the simulation results with $n$ = 5. (E) Bending angle of the bilayer structures under different EL/water compositions with $n$ = 5. (F) Difference of bending angles of the bilayer structures in different environments with $n$ = 5. The data are the differences between the labeled environments with respect to 40\% water. (G) Qualitative simulation and experimental results of the doubled-bilayer design for end-effectors for Type 2 microrobots. Error bars in (E) and (F) are represented with shaded areas with SD. }\label{fig_modular2}
\end{figure}

The Type 2 microrobot has a double-bilayer structure to achieve open and closed configurations. Before applying the printing parameters to print the end-effectors, some general tests with a single bilayer structure have been done to assist the designs. The bending behavior of the bilayer structures is tested with different thickness ratios of responsive (soft) to non-responsive (hard) layers. The 3D design of the suspended bilayer structure is provided in \textbf{Figure \ref{fig_modular2}}A with a hard layer with dimensions of 59 $\micro$m $\times$ 6 $\micro$m $\times$ 20 $\micro$m. The bilayer structures are extruded and suspended, to minimize adhesion for bending, from a fixed base. The bending angle is defined as $\theta$. A qualitative simulation, presented in \textbf{Figure \ref{fig_modular2}}B, for this bilayer structure is also performed to verify the experimental phenomenon. \textbf{Figure \ref{fig_modular2}}C shows the bending results of bilayer structures printed with different soft-to-hard thickness ratios in different water/EL compositions. The soft layers are printed with an LP of 12 mW and an SS of 8 mm/s while the hard layers are printed with an LP of 27.5 mW and a SS of 10 mm/s. It can be seen that the bending angles of the suspended bilayers vary for different thickness ratios and chemical environments. The bending angles with different thickness ratios are also studied with simulation and are presented in \textbf{Figure \ref{fig_modular2}}D showing a peak bending angle for a soft-to-hard thickness ratio of 0.5. With a smaller thickness ratio than this, the soft layer is too thin to generate enough stiffness mismatch to obtain a significant bending. Analytical analysis using the modified Timoshenko bimorph beam theory also proves the existence of the peak~\cite{Huang2016}. The radii can be calculated as 

\begin{equation}
    R = \frac{(h_1+h_2)(8(1+m)^2+(1+mn)(m^2+\frac{1}{mn}))}{6\epsilon(1+m)^2}
\end{equation}

\noindent where $n$ is the ratio of elastic modulus of the hard and soft layers, $m$ is the thickness of the hard and soft layers, and $\epsilon$ is the difference of expansion coefficient of those two layers. Using $n = 2$ and $\epsilon = 1$ will give a local minimum radius at a soft-to-hard ratio corresponding to 0.47. For a certain length of the bilayer structure, the smaller the radius the larger the bending angle, which is the peak. Experimental measurements are provided in \textbf{Figure \ref{fig_modular2}}E. As can be seen, the bending angles decreased as the water content of the environment increased up to 40\%. When the water content reaches 40\%, the soft responsive structure is restored to its designed shape. Then, as you keep increasing the water content to 100\%, a more significant bending angle emerged, as expected. The largest bending angle appears between the thickness ratio of 1.5 to 2.5 for all compositions. The appearance of the peak is in agreement with the simulation result. \textbf{Figure \ref{fig_modular2}}F gives the angle differences between 100\% water and 40\% water and between 0\% water and 40\% water, where 40\% water is a reference state that gives a near zero bending angle. Using the compositions providing the largest (pure water) and smallest (40\% water) bending angles, the angle difference can be as high as 27$^\circ$ with a thickness ratio of $\sim$2. 

After getting the bending result from the suspending bilayer structures, the obtained optimal thickness ratio of 2 is adopted for the end-effector design. \textbf{Figure \ref{fig_modular2}}G shows the printed preliminary design of an end-effector with two bilayer structures working simultaneously like a gripper, verified by a qualitative simulation as well. This gripper design is used as a locking mechanism with the magnetic base in the Type 2 design. The end-effector stays in its designed shape when it is in an environment with 40\% of water so that it can successfully lock the base. When the end-effector is in pure water, the bilayer structures bend and provide a significant opening to disengage the base.

\begin{figure}[t]%
\centering
\includegraphics[width=0.8\linewidth]{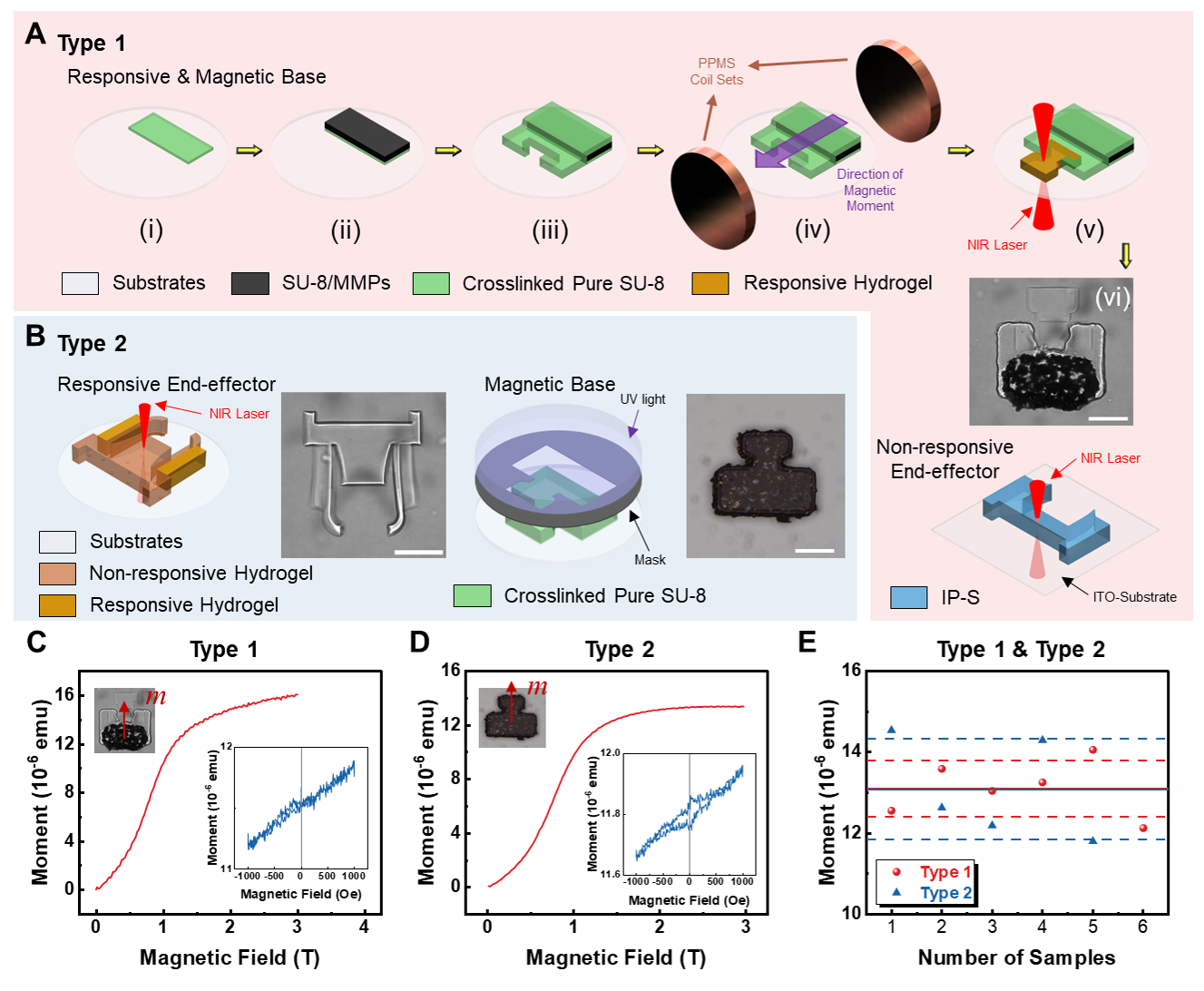}
\caption{Fabrication of different types of modular microrobots. (A) Fabrication process of the magnetic base for Type 1 microrobots containing an RMC and a non-responsive end-effector. (B) Fabrication processes of the Type 2 design with a responsive end-effector and magnetic base. (C) Magnetization of the magnetic base of a Type 1 microrobot. (D) Magnetization of the magnetic base of a Type 2 microrobot. Red curves in (C) and (D) are moments measured during the magnetizations, while the blue curves in the insets are remanence moments measured with four quadrants with a field of 1000 Oe. (E) Remanent moments of multiple samples of the Type 1 and Type 2 magnetic bases. The solid lines represent the average values and the dash lines show the band around the averages with one $\sigma$ (SD). All scale bars: 50 $\micro$m. }\label{fig_fab}
\end{figure}

\subsection{Fabrication of the Modular Microrobots}

The fabrication processes of the modular microrobots are shown in \textbf{Figure \ref{fig_fab}} which combine the traditional photolithography and the TPP technique. The Type 1 responsive and magnetic base (RMB) is fabricated via the route depicted in \textbf{Figure \ref{fig_fab}}A. First, (i) a rectangular pure SU-8 layer is patterned on a circular coverslip with a diameter of 30 mm for better adhesion for the next layer. Then, (ii) a rectangular magnetic piece is photolithographed with a SU-8/magnetic microparticles (MMPs) mixture followed by (iii) another larger layer of pure SU-8 covering the magnetic piece with an additional protruding part for hydrogel printing. The SU-8/MMPs mixture is prepared with a 1 to 1 weight ratio of both pure SU-8 and MMPs followed by a high-speed mixing at 7000 rpm. The second layer of pure SU-8 is designed with an opening for hydrogel printing inside the base to avoid out-of-plane bending along the connection line between the soft and hard materials. Before printing the hydrogel RMC, the base is transferred to a physical properties measurement system (PPMS, Quantum Design Dynacool) to be magnetized under a 3 T field (iv); otherwise, the base shows no significant magnetic moment to be controlled under a magnetic field. This is because magnetic microparticles have no or little apparent magnetic moment before the magnetization since the particles are randomly mixed. The application of a 3 T magnetic field is able to pull the magnetic moment of each microparticle to an identical direction resulting in a significant moment. The PNIPAM hydrogel is then printed using the TPP system with a 63$\times$ oil-immersion objective with an LP of 12 mW and an SS of 8 mm/s (v) using the oil-immersion mode. After the printing process, the Type1 RMB is developed with IPA for an hour to remove any residue. An example of the RMB is shown in \textbf{Figure \ref{fig_fab}}A (vi). The non-responsive end-effectors for the responsive base are simply printed by the TPP process using the 25$\times$ objective with the dip-in mode using the two-photon resist IP-S on an indium-tin-oxide (ITO)-coated substrate. Both the resist and the substrate are obtained from the same company as the TPP system. 

The fabrication of the Type 2 inversed design is simpler with the responsive end-effector achieved by the TPP process while the magnetic base is obtained by photolithography, as shown in \textbf{Figure \ref{fig_fab}}B. The responsive end-effector has two bilayer regions resulting in an opening and closing motion. The printing parameters of the bilayers are the same as shown in \textbf{Figure \ref{fig_modular2}} with an LP of 12 mW and an SS of 8 mm/s for soft regions and an LP of 27.5 mW and an SS of 10 mm/s for the hard regions of the end-effector. The magnetic base is achieved by photolithography of three layers of SU-8, SU-8/MMPs, and SU-8, respectively, using the same mask design. The magnetic base is also magnetized by the PPMS with a magnetic field of 3 T. Both the base and end-effector for the two types of modular microrobots are designed with a width of 120 $\micro$m for an overall footprint size of 120 $\micro$m $\times$ 200 $\micro$m after mating. All features of the microrobots are printed with a thickness of 20 $\micro$m and the parts that are obtained by photolithography have a similar thickness.

\textbf{Figure \ref{fig_fab}}C and \textbf{Figure \ref{fig_fab}}D give the magnetization data of both the bases for Type 1 and Type 2. As can be seen from the red curves, a 3 T magnetic field is enough to achieve saturation. The remanent moments for both bases are tested using a much smaller field (1000 Oe for four quadrants) after the magnetization and the results are presented as insets in \textbf{Figure \ref{fig_fab}}C and \textbf{Figure \ref{fig_fab}}D. As expected, the magnetic base for the Type 1 design has a similar magnetic moment (1.310$\times 10^{-5}$ emu) to the Type 2 (1.308$\times 10^{-5}$ emu) as the areas of the regions containing magnetic materials are similar. Multiple samples were tested for both Type 1 and Type 2 bases and variations in the magnetic moment values were less than 15\%, as shown in \textbf{Figure \ref{fig_fab}}E. 

\begin{figure}[t!]%
\centering
\includegraphics[width=0.8\linewidth]{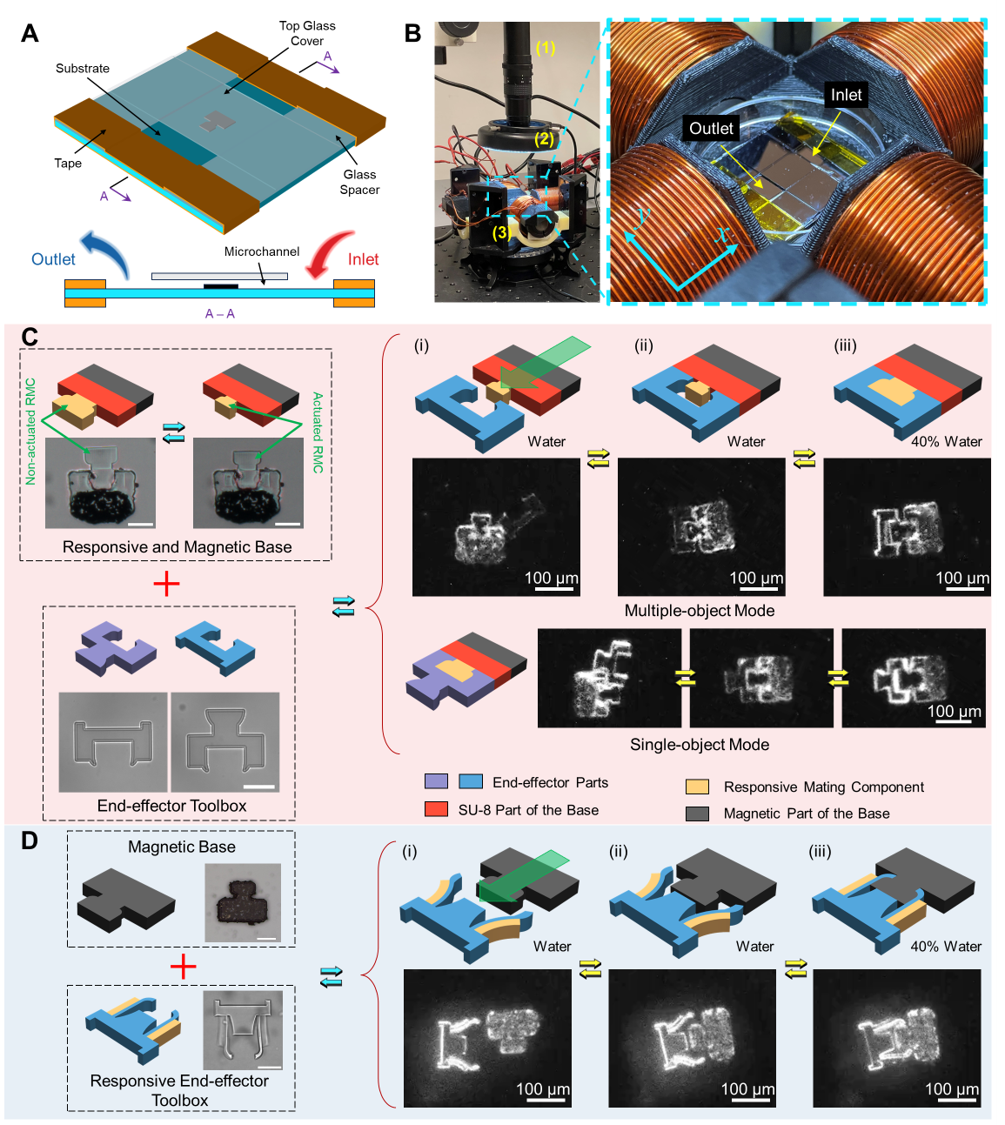}
\caption{Schematic of the modular microrobot with switchable end-effectors. (A) Design of the microchannel used for testing the modular microrobots. (B) Actuation system for generating gradient magnetic field and the prepared microchannel placed in the electromagnetic coils. The microscope, light source, and coils are labeled from 1 to 3, respectively. (C) Mating processes of the Type 1 microrobot with end-effectors for single-pushing and multiple-pushing. (D) Mating process of the Type 2 microrobot. All other scale bars: 50 $\micro$m.}\label{fig_types}
\end{figure}

\subsection{Functional Performance Validation}

The schematic of the concept of operations and experimental setup for the modular microrobots are presented in \textbf{Figure \ref{fig_types}}. The validation experiments are performed in a simple microfluidic environment with a substrate and a coverslip that are spaced by another two coverslips at either end as shown in \textbf{Figure \ref{fig_types}}A. The glass spacers are fixed using tape. Two coverslip spacing layers, each with an ideal thickness of 150 $\micro$m produce a microchannel with a height of 300 $\micro$m. The solvents are exchanged using a pipette by the capillary effect from one side (inlet) to the other side (outlet).  The magnetic bases are controlled using a custom-built electromagnetic coil system capable of generating gradient fields with a setup shown in \textbf{Figure \ref{fig_types}}B. Two pairs of coils are able to generate magnetic gradients along the $x$ and $y$ directions. The prepared microchannel placed in the coil system is also shown in \textbf{Figure \ref{fig_types}}B with the inlet and outlet labeled. The locomotion of the base is controlled with a 3D joystick that is able to produce $x$, $y$, and rotational motions. 

The demonstrations of the use of the Type 1 microrobot are shown in \textbf{Figure \ref{fig_types}}C. Two different types of end-effectors are available for either multiple-object pushing or single-object pushing. Meanwhile, the fabricated RMB is tested for the non-actuated RMC and actuated RMC states using 40\% water and pure water. A significant shrinkage is found as expected. The mating process can be separated into three steps. The first step (i) is where the RMB is disengaged and located away from the end-effector in water. Then the second step (ii) is the mating process where the RMB is controlled in pure water into the mating position. The final step (iii) is when the RMC returns to the non-actuated state and the end-effector and base are locked together as the environment changes from water to 40\% water. The process is the same for both the single-object and multiple-object pushing end-effectors. The illustration of the concept of operations of the Type 2 microrobot is given in \textbf{Figure \ref{fig_types}}D. However, only one type of responsive end-effector is demonstrated. The focus of these tests were on verifying the Type 2 mating mechanism. The overall mating process is similar to Type 1. The difference is in the locking mechanism between the end-effector and the base, achieved via clamping. The responsive end-effector stays open in water during the period when the base is away (i) and moved to the mating-ready position (ii). The mating process is completed by changing the solvent to 40\% water to trigger the closing motion (iii). 

\begin{figure}[h!]%
\centering
\includegraphics[width=0.8\linewidth]{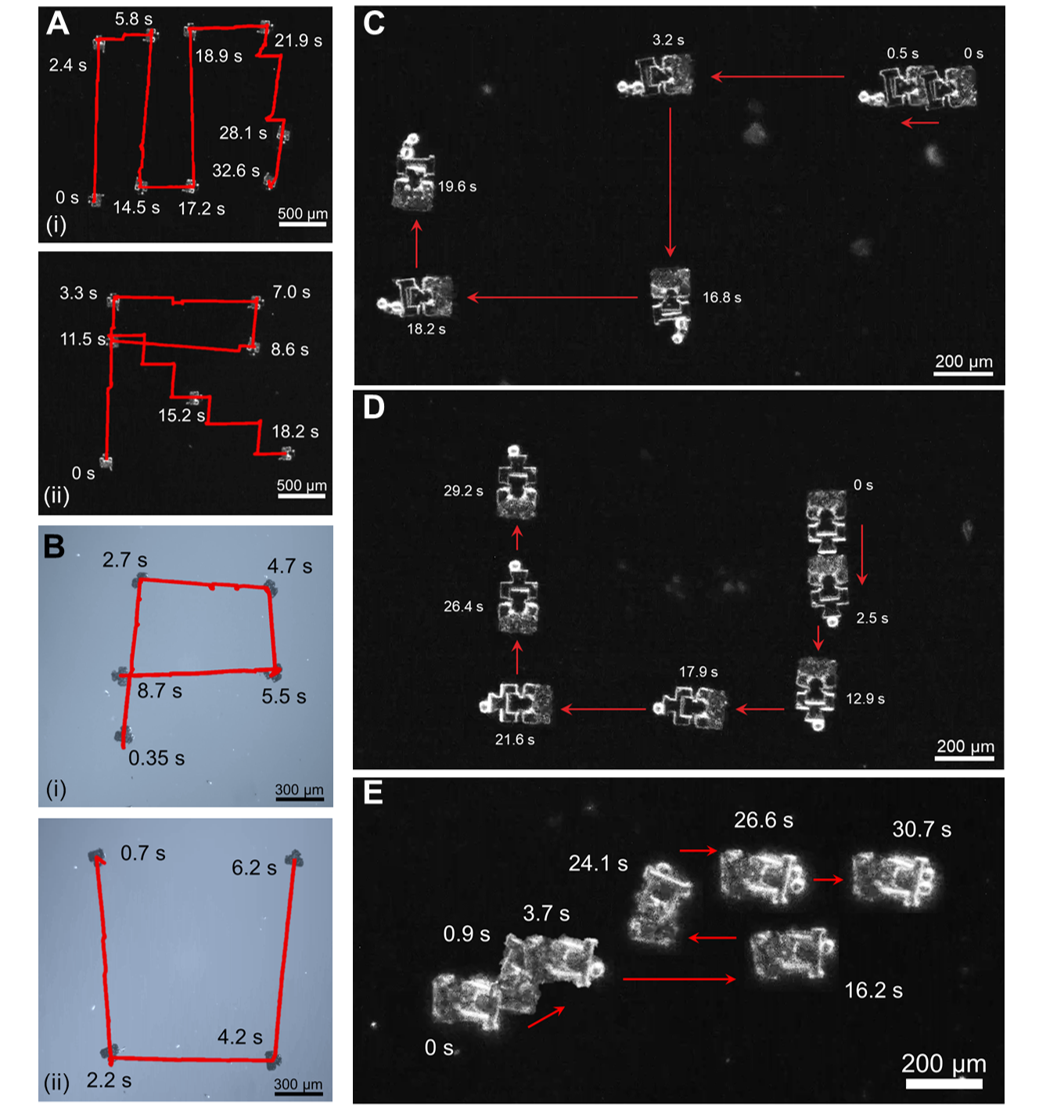}
\caption{Locomotion tests and micromanipulation tasks for different types of modular microrobots. (A) Controlled trajectories of the letters M and R for MicroRobot with a Type 1 magnetic base under the environment of 40\% of water. (B) Controlled trajectories of the letters P and U for Purdue University with a Type 2 magnetic base in water. (C) Manipulation of a sphere cluster using the Type 1 microrobot with an end-effector for multiple-object pushing. (D) Manipulation of a sphere using the Type 1 microrobot with an end-effector for single-object pushing. (E) Manipulation of multiple spheres at different time frames using the Type 2 microrobot with an end-effector for multiple-object pushing.}\label{fig_manip}
\end{figure}

\subsection{Micromanipulation}

After successfully mating their magnetic bases to a particular end-effector, the modular microrobots can be used for micromanipulation tasks. First, controlled propulsion along prescribed trajectories of the magnetic bases is investigated manually. \textbf{Figure \ref{fig_manip}}A shows the controlled experimental trajectories achieving letters of M (i) and R (ii), which present the abbreviation of MicroRobot, using a Type 1 base in 40\% water. The Type 1 base has a RMC made from hydrogels, while the Type 2 base does not. As a result, it is not easy for the Type 1 base to move in 100\% water. A slow speed or a choppy motion is observed because of strong adhesion from the RMC and surface tension in water. However, since a pure water environment is only needed during the mating process and the manipulation tasks are achieved in 40\% of water, the controlled locomotion tests are performed in an environment of 40\% water. The mating process in water can be realized using rotational motions because rotational motions are more efficient. Due to the limitation of the control system, off axes or diagonal motions are achieved by shorter motions decomposed into $x$ and $y$ directions, as can be seen in \textbf{Figure \ref{fig_manip}}A (ii). Any manufacturing defects or adhesive spots with the substrate will give rise to disturbances in the trajectory. The Type 2 base is able to move in 100\% water since the base has a stronger magnetic moment and does have a hydrogel component that could introduce possible substrate adhesion. Comparing the time frames in \textbf{Figure \ref{fig_manip}}A and \textbf{Figure \ref{fig_manip}}B, the Type 1 base shows a faster speed to reach targeted locations even in water. The letters P and U represent Purdue University are drawn with the trajectories in \textbf{Figure \ref{fig_manip}}B (i) and \textbf{Figure \ref{fig_manip}}B (ii), respectively. However, even though both types of the microrobot can move on silicon or glass substrates, a silicon substrate is used under the glass substrate since the end-effectors for both types and the RMC of the Type 1 base are transparent. Therefore, the dark silicon background under a microscope provides better imaging results with improved contrast.

Micromanipulation tasks achieved by Type 1 and Type 2 modular microrobots are demonstrated in \textbf{Figure \ref{fig_manip}}C to \textbf{Figure \ref{fig_manip}}E. \textbf{Figure \ref{fig_manip}}C showcases the Type 1 microrobot with a multiple-object pushing end-effector pushing a sphere cluster. \textbf{Figure \ref{fig_manip}}D presents the Type 1 microrobot manipulating a single 30 $\micro$m diameter sphere using the single-object pushing end-effector. The pushing of the sphere cluster proves that the Type 1 RMB is strong enough to move both the attached end-effector and multiple spheres. \textbf{Figure \ref{fig_manip}}E demonstrations of the use of a Type 2 microrobot the multiple-object pushing end-effector to push multiple spheres sequentially. The Type 2 microrobot successfully picks up sphere 1 at the timeframe $t=$ 3.7 s and transports it until $t=$ 24.1 s. It then picks up sphere 2 at $t=$ 26.6 s and transports it until $t=$ 30.7 s. After manipulating the spheres to the target position, they can be released by applying a simultaneous backward and rotational motion. Using this technique, the objects are typically released roughly within one body-length of the robot from the target position. 

\begin{figure}[t]%
\centering
\includegraphics[width=0.8\linewidth]{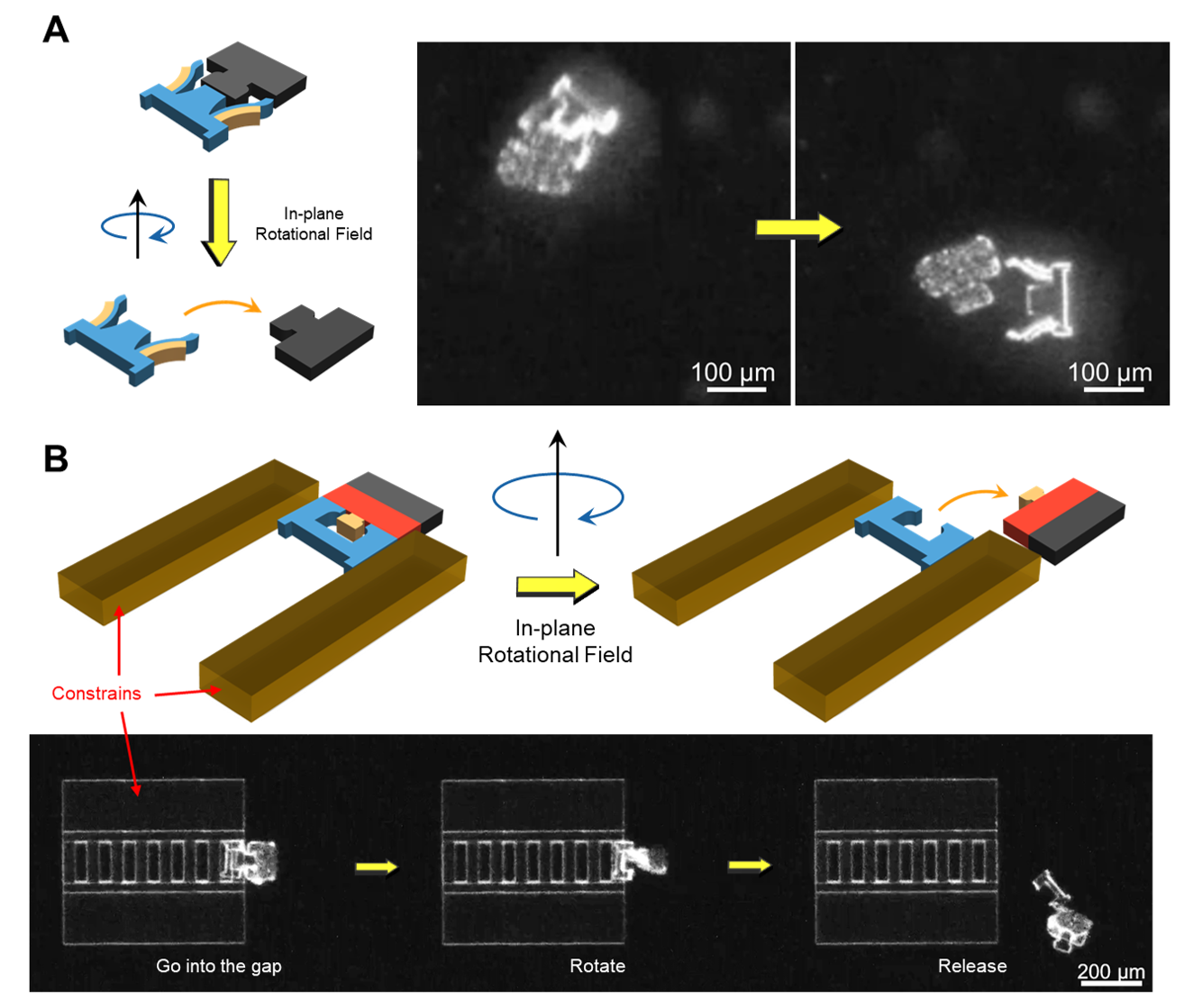}
\caption{Detachment of end-effectors of the modular microrobots. (A) Detachment of the Type 2 modular microrobot using a rotational field in water. (B) Detachment of the Type 1 modular microrobot using a rotational field with the assistance of a channeled constraint in water.}\label{fig_det}
\end{figure}

\subsection{End-effector Detachment}

We have demonstrated the mating process and micromanipulation with the two different types of modular microrobots. However, it is essential to prove that the magnetic base can be detached from the end-effector for end-effector exchange. \textbf{Figure \ref{fig_det}} illustrates the detachment process for both types of microrobots. The Type 2 microrobot is presented first in \textbf{Figure \ref{fig_det}}A. The base can be detached easily from the end-effector as long as the double-bilayer structure is open in water by using a rotational motion. This is because when stimulated, the double-bilayer structure produces a wide opening around the male feature of the base. The situation for the Type 1 microrobot is different. Since the male RMC becomes smaller due to shrinkage once stimulated, the RMC is still surrounded by the female feature of the end-effector with a small tolerance gap. This results in a strong surface tension between the components, especially in 100\% water. Therefore, this surface tension and inherent adhesion between the end-effector and the base prevent the detachment using a simple rotational motion. Therefore, assistance is required for the detachment of the Type 1 microrobot. As can be seen in \textbf{Figure \ref{fig_det}}B, two large rectangular pieces are printed on a silicon substrate creating the microchannel. The gap between the two pieces is used to provide a constraint for the end-effectors. A top enclosure is also introduced to prevent any possible out-of-plane motions during the detachment. A rotational motion is performed after moving the end-effector inside the gap. Once the rotational motion of the end-effector is constrained by the walls, the end-effector and the base can be successfully detached via rotational motion of the RMB. 

\section{Conclusion}

In this paper, the design and fabrication of two types of modular microrobots are proposed by using both traditional photolithography and two-photon polymerization 3D printing techniques. The magnetic base and the end-effector of the modular microrobot are connected with a male-female design which is realized through the use of stimuli-responsive hydrogels (responsive mating components). Two types of modular microrobots based on the location of responsive mating components are discussed. The mating process of the magnetic bases and the end-effectors for both types are successfully validated in different solvents that trigger the process. Different micromanipulation tasks using the proposed modular microrobots are demonstrated with both types of microrobots. The detachment process of the magnetic bases from the end-effectors is also demonstrated. The paper has successfully fabricated and demonstrated the modular microrobot with an overall footprint of 120 $\micro$m $\times$ 200 $\micro$m that is able to perform different tasks with the same magnetic base by using different end-effectors. 


\medskip
\textbf{Supporting Information} \par 
Supporting Information is available from the Wiley Online Library or from the author.

\medskip
\textbf{Acknowledgements} \par 
This work was supported by the National Science Foundation (NSF IIS Award 1763689 and NSF CMMI Award 2018570) and Purdue Institute for Cancer Research DDMS Program Special Projects Award 22-23. The authors would like to thank Prof. Adrian Buganza Tepole in the School of Mechanical Engineering at Purdue University for providing access to ABAQUS. The authors would also like to thank Yang Yang from the School of Mechanical Engineering at Purdue University for helping with part of the photolithography process. 

\medskip

%
\bibliographystyle{MSP}
\bibliography{references}




\begin{figure}[h]
  \includegraphics{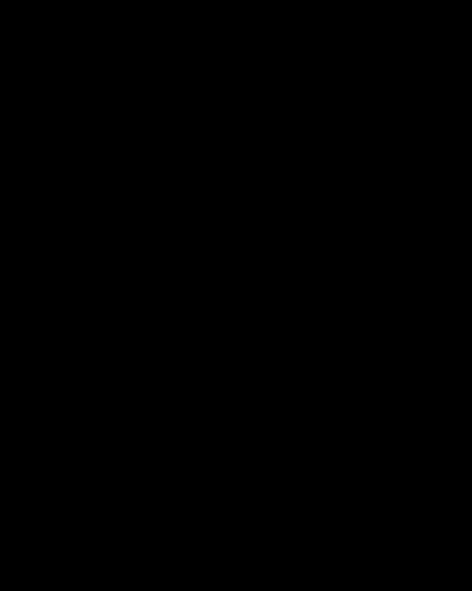}
  \caption*{Biography}
\end{figure}


\begin{figure}[h]
\textbf{Table of Contents}\\
\medskip
  \includegraphics{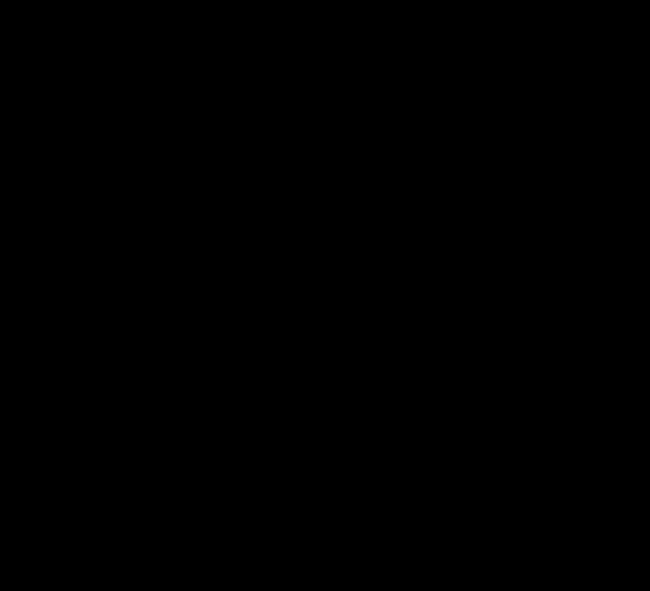}
  \medskip
  \caption*{ToC Entry}
\end{figure}

\end{document}